\documentclass[sigconf]{acmart}

\newcommand{\statute}[1]{\begin{quote} \emph{#1} \end{quote}}
\newcommand{\mysection}[1]{\vspace{-5pt}\section{#1}}
\newcommand{\mysubsection}[1]{\vspace{-5pt}
\subsection{#1}}
\usepackage{graphicx,wrapfig,multirow}
\usepackage{subfigure}
\usepackage[linesnumbered,ruled,vlined]{algorithm2e}

\AtBeginDocument{%
  \providecommand\BibTeX{{%
    \normalfont B\kern-0.5em{\scshape i\kern-0.25em b}\kern-0.8em\TeX}}}

\begin{document}


\title{A Dataset for Statutory Reasoning in Tax Law Entailment and Question Answering}

\author{Nils Holzenberger}
\affiliation{%
  \institution{Johns Hopkins University}
  \streetaddress{3400 N Charles Street}
  \city{Baltimore}
  \state{Maryland}
  \postcode{21218}
  \country{USA}}
\email{nilsh@jhu.edu}

\author{Andrew Blair-Stanek}
\affiliation{%
  \institution{U. of Maryland Carey School of Law}
  \streetaddress{500 W. Baltimore St}
  \city{Baltimore}
  \state{Maryland}
  \postcode{21201}
  \country{USA}}
\affiliation{%
  \institution{Johns Hopkins University}
  \streetaddress{3400 N Charles Street}
  \city{Baltimore}
  \state{Maryland}
  \postcode{21218}
  \country{USA}}
\email{ablair-stanek@law.umaryland.edu}

\author{Benjamin Van Durme}
\affiliation{%
  \institution{Johns Hopkins University}
  \streetaddress{3400 N Charles Street}
  \city{Baltimore}
  \state{Maryland}
  \postcode{21218}
  \country{USA}}
\email{vandurme@cs.jhu.edu}

\renewcommand{\shortauthors}{Holzenberger, Blair-Stanek and Van Durme}

\begin{abstract}
 Legislation can be viewed as a body of prescriptive rules expressed in natural language. The application of legislation to facts of a case we refer to as statutory reasoning, where those facts are also expressed in natural language.  Computational statutory reasoning is distinct from most existing work in machine reading, in that much of the information needed for deciding a case is declared exactly once (a law), while the information needed in much of machine reading tends to be learned through distributional language statistics.  To investigate the performance of natural language understanding approaches on statutory reasoning, we introduce a dataset, together with a legal-domain text corpus. Straightforward application of machine reading models exhibits low out-of-the-box performance on our questions, whether or not they have been fine-tuned to the legal domain. We contrast this with a hand-constructed Prolog-based system, designed to fully solve the task.  These experiments support a discussion of the challenges facing statutory reasoning moving forward, which we argue is an interesting real-world task that can motivate the development of models able to utilize prescriptive rules specified in natural language.

\end{abstract}

\ccsdesc[500]{Applied computing~Law}
\ccsdesc[300]{Computing methodologies~Natural language processing}
\ccsdesc[300]{Computing methodologies~Knowledge representation and reasoning}

\keywords{Law, NLP, Reasoning, Prolog}


\maketitle

\mysection{Introduction}

Early artificial intelligence research focused on highly-performant, narrow-domain reasoning models, for instance in health \citep{ledley1959reasoning,shortliffe1975model,miller1982internist} and law \citep{mccarty1976reflections,hellawell1980computer}. Such \emph{expert systems} relied on hand-crafted inference rules and domain knowledge, expressed and stored with the formalisms provided by databases \citep{feigenbaum1992expert}. The main bottleneck of this approach is that experts are slow in building such knowledge bases and exhibit imperfect recall, which motivated research into models for automatic information extraction (e.g. \citet{lafferty2001conditional}). Systems for large-scale automatic knowledge base construction have improved (e.g. \citet{etzioni2008open,mitchell2018never}), as well as systems for sentence level semantic parsing \citep{zhang2019broad}. Among others, this effort has led to question-answering systems for games \citep{ferrucci2010building} and, more recently, for science exams \citep{friedland2004project,gunning2010project,clark2019f}. The challenges include extracting ungrounded knowledge from semi-structured sources, e.g. textbooks, and connecting high-performance symbolic solvers with large-scale language models.

In parallel, models have begun to consider task definitions like Machine Reading (MR)~\citep{rajpurkar2018know} and Recognizing Textual Entailment (RTE)~\citep{cooper1996using,dagan2005pascal} as not requiring the use of explicit structure. Instead, the problem is cast as one of mapping inputs to high-dimensional, dense representations that implicitly encode meaning~\citep{devlin2018bert,raffel2019exploring}, and are employed in building classifiers or text decoders, bypassing classic approaches to symbolic inference.

This work is concerned with the problem of statutory reasoning \citep{yoshioka2018overview, zhong2019jec}: how to reason about an example situation, a \emph{case}, based on complex rules provided in natural language. In addition to the reasoning aspect, we are motivated by the lack of contemporary systems to suggest legal opinions: while there exist tools to aid lawyers in retrieving relevant documents for a given case, we are unaware of any strong capabilities in automatic statutory reasoning.

\begin{figure}[t]
    \centering
        \includegraphics[width=\columnwidth]{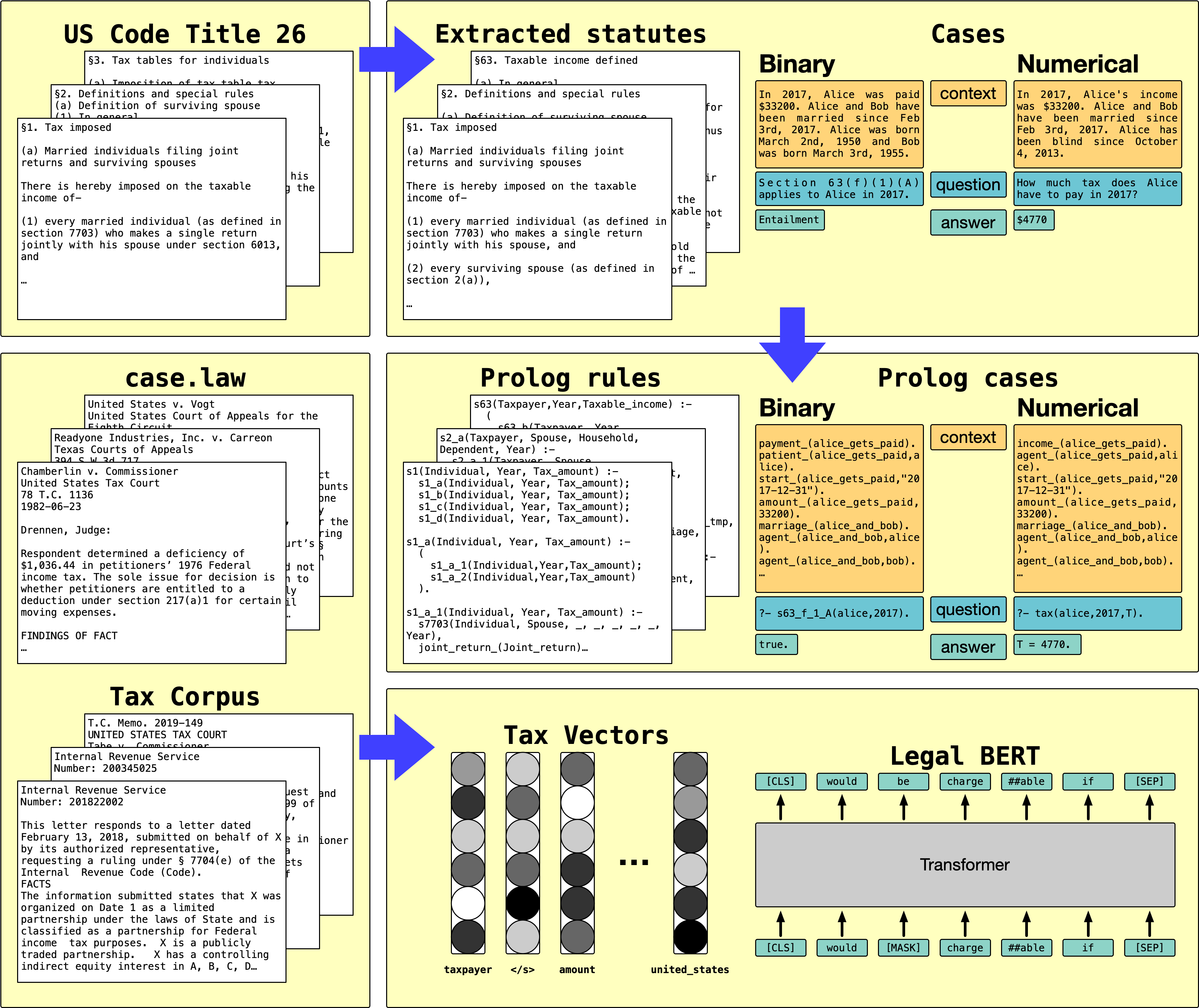}
    \caption{Resources. Corpora on the left hand side were used to build the datasets and models on the right hand side.}
    \label{fig:resources}
    \vspace{-15pt}
\end{figure}

Our contributions, summarized in Figure \ref{fig:resources}, include a novel dataset based on US tax law, together with test cases (Section \ref{sec:dataset}). Decades-old work in expert systems could solve problems of the sort we construct here, based on manually derived rules: we replicate that approach in a Prolog-based system that achieves 100\% accuracy on our examples (Section \ref{sec:prolog}). Our results demonstrate that straightforward application of contemporary Machine Reading models is not sufficient for our challenge examples (Section \ref{sec:experiments}), whether or not they were adapted to the legal domain (Section \ref{sec:legal}). This is meant to provoke the question of whether we should be concerned with: (a) improving methods in semantic parsing in order to replace manual transduction into symbolic form; or (b) improving machine reading methods in order to avoid explicit symbolic solvers. We view this work as part of the conversation including recent work in multi-hop inference \citep{yang2018hotpotqa}, where our task is more domain-specific but potentially more challenging.

\begin{figure}[t]
    \centering
    \includegraphics[width=\columnwidth]{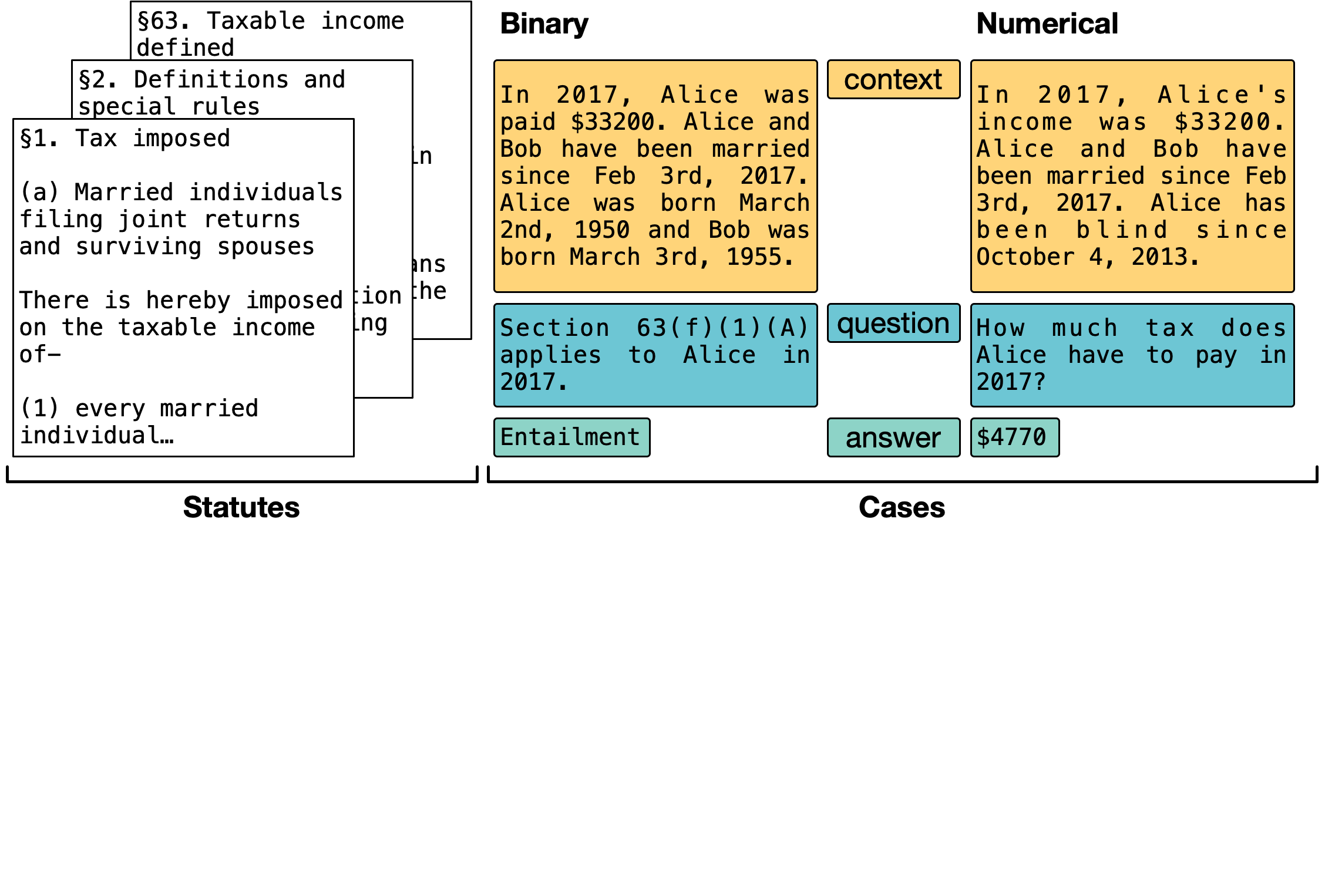}
    \vspace{-75pt}
    \caption{Sample cases from our dataset. The questions can be answered by applying the rules contained in the statutes to the context.}
    \label{fig:task}
    \vspace{-10pt}
\end{figure}

\mysection{Dataset}
\label{sec:dataset}

Here, we describe our main contribution, the StAtutory Reasoning Assessment dataset (SARA): a set of rules extracted from the statutes of the US Internal Revenue Code (IRC), together with a set of natural language questions which may only be answered correctly by referring to the rules\footnote{The dataset can be found under \url{https://nlp.jhu.edu/law/}}.

The IRC\footnote{\url{https://uscode.house.gov/browse/prelim@title26&edition=prelim}} contains rules and definitions for the imposition and calculation of taxes. It is subdvided into sections, which in general, define one or more terms: section 3306 defines the terms employment, employer and wages, for purposes of the federal unemployment tax.  Sections are typically structured around a general rule, followed by a number of exceptions. Each section and its subsections may be cast as a predicate whose truth value can be checked against a state of the world. For instance, subsection 7703(a)(2): \statute{an individual legally separated from his spouse under a decree of divorce or of separate maintenance shall not be considered as married} can be checked given an individual.

Slots are another major feature of the law. Each subsection refers to a certain number of slots, which may be filled by existing entities (in the above, \emph{individual}, \emph{spouse}, and \emph{decree of divorce or of separate maintenance}). Certain slots are implicitly filled: \textsection 7703(a)(1) and (b)(3) mention a ``spouse", which must exist since the ``individual" is married. Similarly, slots which have been filled earlier in the section may be referred to later on. For instance, ``household" is mentioned for the first time in \textsection 7703(b)(1), then again in \textsection 7703(b)(2)
and in \textsection 7703(b)(3). Correctly resolving slots is a key point in successfully applying the law.

Overall, the IRC can be framed as a set of predicates formulated in human language. The language used to express the law has an open texture~\citep{hart2012concept}, which makes it particularly challenging for a computer-based system to determine whether a subsection applies, and to identify and fill the slots mentioned. This makes the IRC an excellent corpus to build systems that reason with rules specified in natural language, and have good language understanding capabilities.

\mysubsection{Statutes and test cases}

As the basis of our set of rules, we selected sections of the IRC well-supported by Treasury Regulations, covering tax on individuals (\textsection 1), marriage and other legal statuses (\textsection 2, 7703), dependents (\textsection 152), tax exemptions and deductions (\textsection 63, 68, 151) and employment (\textsection 3301, 3306). We simplified the sections to (1) remove highly specific sections (e.g. those concerning the employment of sailors) in order to keep the statutes to a manageable size, and (2) ensure that the sections only refer to sections from the selected subset. For ease of comparison with the original statutes, we kept the original numbering and lettering, with no adjustment for removed sections. For example, there is a section 63(d) and a section 63(f), but no section 63(e). We assumed that any taxable year starts and ends at the same time as the corresponding calendar year.

For each subsection extracted from the statutes, we manually created two paragraphs in natural language describing a case, one where the statute applies, and one where it does not. These snippets, formulated as a logical entailment task, are meant to test a system's understanding of the statutes, as illustrated in Figure \ref{fig:task}. The cases were vetted by a law professor for coherence and plausibility. For the purposes of machine learning, the cases were split into 176 train and 100 test samples, such that (1) each pair of positive and negative cases belongs to the same split, and (2) each section is split between train and test in the same proportions as the overall split.

Since tax legislation makes it possible to predict how much tax a person owes, we created an additional set of 100 cases where the task is to predict how much tax someone owes. Those cases were created by randomly mixing and matching pairs of cases from the first set of cases, and resolving inconsistencies manually. Those cases are no longer a binary prediction task, but a task of predicting an integer. The prediction results from taking into account the entirety of the statutes, and involves basic arithmetic. The 100 cases were randomly split into 80 training and 20 test samples.

Because the statutes were simplified, the answers to the cases are not those that would be obtained with the current version of the IRC. Some of the IRC counterparts of the statutes in our dataset have been repealed, amended, or adjusted to reflect inflation.

\mysubsection{Key features of the corpus}
\label{subsec:key}

While the corpus is based on a simplification of the Internal Revenue Code, care was taken to retain prominent features of US law. We note that the present task is only one aspect of legal reasoning, which in general involves many more modes of reasoning, in particular interpreting regulations and prior judicial decisions. The following features are quantified in Tables \ref{tab:corpus_statistics_cross_refs} to \ref{tab:corpus_statistics_numerical}.

\emph{Reasoning with time.} The timing of events (marriage, retirement, income...) is highly relevant to determining whether certain sections apply, as tax is paid yearly. In total, 62 sections refer to time. Some sections require counting days, as in \textsection 7703(b)(1): \statute{a household which constitutes for more than one-half of the taxable year the principal place of abode of a child} or taking into account the absolute point in time as in \textsection 63(c)(7): \statute{In the case of a taxable year beginning after December 31, 2017, and before January 1, 2026-}

\emph{Exceptions and substitutions.} Typically, each section of the IRC starts by defining a general case and then enumerates a number of exceptions to the rule. Additionally, some rules involve applying a rule after substituting terms. A total of 50 sections formulate an exception or a substitution. As an example, \textsection 63(f)(3): \statute{In the case of an individual who is not married and is not a surviving spouse, paragraphs (1) and (2) shall be applied by substituting ``\$750" for ``\$600".}

\emph{Numerical reasoning.} Computing tax owed requires knowledge of the basic arithmetic operations of adding, subtracting, multiplying, dividing, rounding and comparing numbers. 55 sections involve numerical reasoning. The operation to be used needs to be parsed out of natural text, as in \textsection 1(c)(2): \statute{\$3,315, plus 28\% of the excess over \$22,100 if the taxable income is over \$22,100 but not over \$53,500}

\emph{Cross-references.} Each section of the IRC will typically reference other sections. Table \ref{tab:corpus_statistics_cross_refs} shows how this feature was preserved in our dataset. There are explicit references within the same section, as in \textsection 7703(b)(1): \statute{an individual who is married (within the meaning of subsection (a)) and who files a separate return} explicit references to another section, as in \textsection 3301: \statute{There is hereby imposed on every employer (as defined in section 3306(a)) for each calendar year an excise tax} and implicit references, as in \textsection 151(a), where ``taxable income" is defined in \textsection 63: \statute{the exemptions provided by this section shall be allowed as deductions in computing taxable income.}

\begin{table}
  \small
  \begin{tabular}{l|rr}
    Cross-references & Explicit & Implicit \\
    \hline Within the section & 30 & 25 \\
    To another section & 34 & 44 \\
  \end{tabular}
  \caption{Number of subsections containing cross-references}
  \label{tab:corpus_statistics_cross_refs}
  \vspace{-20pt}
\end{table}

\emph{Common sense knowledge.} Four concepts, other than time, are left undefined in our statutes: (1) kinship, (2) the fact that a marriage ends if either spouse dies, (3) if an event has not ended, then it is ongoing; if an event has no start, it has been true at any time before it ends; and some events are instantaneous (e.g. payments ), (4) a person's gross income is the sum of all income and payments received by that person.

\emph{Hierarchical structure.} Law statutes are divided into sections, themselves divided into subsections, with highly variable depth and structure. This can be represented by a tree, with a special ROOT node of depth 0 connecting all the sections. This tree contains 132 leaves and 193 nodes (node includes leaves). Statistics about depth are in Table \ref{tab:corpus_statistics_structure}.

\begin{table}
  \small
    \begin{tabular}{l|rrrr}
     & min & max & avg $\pm$ stddev & median \\
    \hline Depth of leaf & 1 & 6 & 3.6 $\pm$ 0.8 & 4 \\
    Depth of node & 0 & 6 & 3.2 $\pm$ 1.0 & 3 \\
  \end{tabular}
\caption{Statistics about the tree structure of the statutes}
\label{tab:corpus_statistics_structure}
\vspace{-20pt}
\end{table}

\begin{table}
  \small
  \begin{tabular}{lr|rrrrr}
    \hline Vocabulary & & & train & 867 & statutes & 768 \\
     size & & & test & 535 & combined & 1596 \\
    \hline & & min & max & avg & stddev & median \\
    \hline Sentence  & train & 4 & 138 & 12.3 & 9.1 & 11 \\
    length & test & 4 & 34 & 11.6 & 4.5 & 10 \\
    (in words) & statutes & 1 & 88 & 16.5 & 14.9 & 12.5 \\
    & combined & 1 & 138 & 12.7 & 9.5 & 11 \\
    \hline Case length & train & 1 & 9 & 4.2 & 1.7 & 4 \\
    (in & test & 2 & 7 & 3.8 & 1.3 & 4 \\
    sentences) & combined & 1 & 9 & 4.1 & 1.6 & 4 \\
    \hline Case & train & 17 & 179 & 48.5 & 22.2 & 43 \\
    length & test & 17 & 81 & 41.6 & 14.7 & 38 \\
    (in words) & combined & 17 & 179 & 46.3 & 20.3 & 41 \\
    \hline Section & sentences & 2 & 16 & 8.3 & 4.7 & 9 \\
    length & words & 62 & 1151 & 488.9 & 310.4 & 549 \\
    \hline
  \end{tabular}
\caption{Language statistics. The word ``combined'' means merging the corpora mentioned above it.}
\label{tab:corpus_statistics_language}
\vspace{-20pt}
\end{table}

\begin{table}
  \small
  \begin{tabular}{rrrrrrr}
    & min & max & average & stddev & median \\
    \hline train & 0 & 2,242,833 & 85,804.86 & 258,179.30 & 15,506.50 \\
    test & 0 & 243,097 & 65,246.50 & 78,123.13 & 26,874.00 \\
    combined & 0 & 2,242,833 & 81,693.19 & 233,695.33 & 17,400.50 \\
\end{tabular}
\caption{Answers to numerical questions (in \$).}
\label{tab:corpus_statistics_numerical}
\vspace{-20pt}
\end{table}

\mysection{Prolog solver}
\label{sec:prolog}

It has been shown that subsets of statutes can be expressed in first-order logic, as described in Section \ref{sec:related}. As a reaffirmation of this, and as a topline for our task, we have manually translated the statutes into Prolog rules and the cases into Prolog facts, such that each case can be answered correctly by a single Prolog query\footnote{The Prolog program can be found under \url{https://nlp.jhu.edu/law/}}. The Prolog rules were developed based on the statutes, meaning that the Prolog code clearly reflects the semantics of the textual form, as in \citet{gunning2010project}. This is primarily meant as a proof that a carefully crafted reasoning engine, with perfect natural language understanding, can solve this dataset. There certainly are other ways of representing this given set of statutes and cases. The point of this dataset is not to design a better Prolog system, but to help the development of language understanding models capable of reasoning.

\mysubsection{Statutes}

Each subsection of the statutes was translated with a single rule, true if the section applies, false otherwise. In addition, subsections define slots that may be filled and reused in other subsections, as described in Section \ref{sec:dataset}. To solve this coreference problem, any term appearing in a subsection and relevant across subsections is turned into an argument of the Prolog rule. The corresponding variable may then be bound during the execution of a rule, and reused in a rule executed later. Unfilled slots correspond to unbound variables.

To check whether a given subsection applies, the Prolog system needs to rely on certain predicates, which directly reflect the facts contained in the natural language descriptions of the cases. For instance, how do we translate \emph{Alice and Bob got married on January 24th, 1993} into code usable by Prolog? We rely on a set of 61 predicates, following neo-davidsonian semantics \citep{davidson-67,casteneda-67,parsons1990events}. The level of detail of these predicates is based on the granularity of the statutes themselves. Anything the statutes do not define, and which is typically expressed with a single word, is potentially such a predicate: marriage, residing somewhere, someone paying someone else, etc. The example above is translated in Figure \ref{tab:facts}.

\begin{wrapfigure}{R}{0pt}
\small
\begin{tabular}{l}
\texttt{marriage\_(alice\_and\_bob).}\\
\texttt{agent\_(alice\_and\_bob, alice).}\\
\texttt{agent\_(alice\_and\_bob, bob).}\\
\texttt{start\_(alice\_and\_bob, "1993-01-24").} 
\end{tabular}
\caption{Example predicates used.}
\vspace{-10pt}
\label{tab:facts}
\end{wrapfigure}

\mysubsection{Cases}

The natural language description of each case was manually translated into the facts mentioned above. The question or logical entailment prompt was translated into a Prolog query. For instance, \emph{Section 7703(b)(3) applies to Alice maintaining her home for the year 2018.} translates to \texttt{s7703\_b\_3(alice,home,2018).} and \emph{How much tax does Alice have to pay in 2017?} translates to \texttt{tax(alice,2017,Amount).}

In the broader context of computational statutory reasoning, the Prolog solver has three limitations. First, producing it requires domain experts, while automatic generation is an open question. Second, translating natural language into facts requires semantic parsing capabilities. Third, small mistakes can lead to catastrophic failure. An orthogonal approach is to replace logical operators and explicit structure with high-dimensional, dense representations and real-valued functions, both learned using distributional statistics. Such a machine learning-based approach can be adapted to new legislation and new domains automatically.

\mysection{Legal NLP}
\label{sec:legal}



As is commonly done in MR, we pretrained our models using two unsupervised learning paradigms on a large corpus of legal text.

\mysubsection{Text corpus}

We curated a corpus consisting solely of freely-available tax law documents with 147M tokens.  The first half is drawn from \citet{caselawproject}, a project of Harvard's Law Library that scanned and OCR'ed many of the library's case-law reporters, making the text available upon request to researchers.  The main challenge in using this resource is that it contains 1.7M U.S. federal cases, only a small percentage of which are on tax law (as opposed to criminal law, breach of contract, bankruptcy, etc.).  Classifying cases by area is a non-trivial problem \citep{soh-classify}, and tax-law cases are litigated in many different courts.  We used the heuristic of classifying a case as being tax-law if it met one of the following criteria:  the Commissioner of Internal Revenue was a party; the case was decided by the U.S. Tax Court; or, the case was decided by any other federal court, other than a trade tribunal, with the United States as a party, and with the word \textit{tax} appearing in the first 400 words of the case's written opinion.  

The second half of this corpus consists of IRS private letter rulings and unpublished U.S. Tax Court cases.  IRS private letter rulings are similar to cases, in that they apply tax law to one taxpayer's facts; they differ from cases in that they are written by IRS attorneys (not judges), have less precedential authority than cases, and redact names to protect taxpayer privacy.  Unpublished U.S. Tax Court cases are viewed by the judges writing them as less important than those worthy of publication.  These were downloaded as PDFs from the IRS and Tax Court websites, OCR'ed with \texttt{tesseract} if needed, and otherwise cleaned.  

\mysubsection{Tax vectors}

Before training a \texttt{word2vec} model \citep{mikolov2013distributed} on this corpus, we did two tax-specific preprocessing steps to ensure that semantic units remained together.  First, we put underscores between multi-token collocations that are tax terms of art, defined in either the tax code, Treasury regulations, or a leading tax-law dictionary.  Thus, ``surviving spouse" became the single token ``surviving\_spouse".  Second, we turned all tax code sections and Treasury regulations into a single token, stripped of references to subsections, subparagraphs, and subclauses. Thus, ``Treas. Reg. \textsection  1.162-21(b)(1)(iv)" became the single token ``sec\_1\_162\_21".  The vectors were trained at 500 dimensions using skip-gram with negative sampling.  A window size of 15 was found to maximize performance on twelve human-constructed analogy tasks.

\mysubsection{Legal BERT}

We performed further training of BERT \citep{devlin2018bert}, on a portion of the full \texttt{case.law} corpus, including both state and federal cases.  We did not limit the training to tax cases.  Rather, the only cases excluded were those under 400 characters (which tend to be summary orders with little semantic content) and those before 1970 (when judicial writing styles had become recognizably modern).  We randomly selected a subset of the remaining cases, and broke all selected cases into chunks of exactly 510 tokens, which is the most BERT's architecture can handle.  Any remaining tokens in a selected case were discarded.  Using solely the masked language model task (i.e. not next sentence prediction), starting from \texttt{Bert-Base-Cased}, we trained on 900M tokens.  

The resulting Legal BERT has the exact same architecture as \texttt{Bert-Base-Cased} but parameters better attuned to legal tasks.  We applied both models to the natural language questions and answers in the corpus we introduce in this paper.  While \texttt{Bert-Base-Cased} had a perplexity of 14.4, Legal BERT had a perplexity of just 2.7, suggesting that the further training on 900M tokens made the model much better adapted to legal queries.  

We also probed how this further training impacted ability to handle fine-tuning on downstream tasks.  The downstream task we chose was identifying legal terms in case texts.  For this task, we defined legal terms as any tokens or multi-token collocations that are defined in \textit{Black's Law Dictionary}~\citep{blackslawdictionary}, the premier legal dictionary.  We split the legal terms into training/dev/test splits.  We put a 4-layer fully-connected MLP on top of both \texttt{Bert-Base-Cased} and Legal BERT, where the training objective was \texttt{B-I-O} tagging of tokens in 510-token sequences.  We trained both on a set of 200M tokens randomly selected from \texttt{case.law} cases not previously seen by the model and not containing any of the legal terms in dev or test, with the training legal terms tagged using string comparisons.  We then tested both fine-tuned models' ability to identify legal terms from the test split in case law.  The model based on \texttt{Bert-Base-Cased} achieved F1 = 0.35, whereas Legal BERT achieved F1 = 0.44. As a baseline, two trained lawyers given the same task on three 510-token sequences each achieved F1 = 0.26. These results indicate that Legal BERT is much better adapted to the legal domain than \texttt{Bert-Base-Cased}. Black's Law Dictionary has well-developed standards for what terms are or are not included. BERT models learn those standards via the train set, whereas lawyers are not necessarily familiar with them. In addition, pre-processing dropped some legal terms that were subsets of too many others, which the lawyers tended to identify. This explains how BERT-based models could outperform trained humans.  

\mysection{Experiments}
\label{sec:experiments}

\vspace{2pt}
\mysubsection{BERT-based models}
\label{subsec:bert}

In the following, we frame our task as textual entailment and numerical regression. A given entailment prompt $q$ mentions the relevant subsection (as in Figure \ref{fig:task})\footnote{The code for these experiments can be found under \url{https://github.com/SgfdDttt/sara}}. We extract $s$, the text of the relevant subsection, from the statutes. In $q$, we replace \emph{Section XYZ applies} with \emph{This applies}. We feed the string ``[CLS] + $s$ + [SEP] + $q$ + $c$ + [SEP]", where ``+" is string concatenation, to BERT \citep{devlin2018bert}. Let $r$ be the vector representation of the token [CLS] in the final layer. The answer (entailment or contradiction) is predicted as $g(\theta_1 \cdot r)$ where $\theta_1$ is a learnable parameter and $g$ is the sigmoid function. For numerical questions, all statutes have to be taken into account, which would exceed BERT's length limit. We encode ``[CLS] all [SEP] + $q$ + $c$ + [SEP]" into $r$ and predict the answer as $\mu + \sigma \theta_2 \cdot r$ where $\theta_2$ is a learned parameter, and $\mu$ and $\sigma$ are the mean and standard deviation of the numerical answers on the training set.

For entailment, we use a cross-entropy loss, and evaluate the models using accuracy. We frame the numerical questions as a taxpayer having to compute tax owed. By analogy with the concept of ``substantial understatement of income tax'' from \textsection 6662(d), we define $\Delta(y,\hat y) = \frac{|y - \hat y|}{\max(0.1y,5000)}$ where $y$ is the true amount of tax owed, and $\hat y$ is the taxpayer's prediction. The case $\Delta(y,\hat y) \geq 1$ corresponds to a substantial over- or understatement of tax. We compute the fraction of predictions $\hat y$ such that $\Delta(y,\hat y) < 1$ and report that as numerical accuracy.\footnote{For a company, a goal would be to have 100\% accuracy (resulting in no tax penalties) while paying the lowest amount of taxes possible (giving them something of an interest-free loan, even if the IRS eventually collects the understated tax).} The loss function used is:

$$\mathcal{L} = \sum_{i \in I_1} y_i \log \hat y_i + (1-y_i) \log (1-\hat y_i) + \sum_{i \in I_2} \max(\Delta(y_i, \hat y_i)-1,0)$$

\noindent where $I_1$ (resp. $I_2$) is the set of entailment (resp. numerical) questions, $y_i$ is the ground truth output, and $\hat y_i$ is the model's output.

\begin{table}
  \small
  \begin{tabular}{l l l r r}
    {\bf Model}  & {\bf Features} & {\bf Inputs} & {\bf Entailment} & {\bf Numerical} \\
    \hline Baseline    & -           & -        & 50 & 20 $\pm$ 15.8  \\
    \hline BERT-       & BERT        & question & 48 & 20 $\pm$ 15.8 \\
    based              &             & context  & 55 & 15 $\pm$ 14.1 \\
                       &             & statutes & 49 & 5 $\pm$ 8.6 \\
                       & + unfreeze  & statutes & 53 & 20 $\pm$ 15.8    \\
                       & Legal BERT  & question & 48 & 5 $\pm$ 8.6 \\
                       &             & context  & 49 & 5 $\pm$ 8.6 \\
                       &             & statutes & 48 & 5 $\pm$ 8.6 \\
                       & + unfreeze  & statutes & 49 & 15 $\pm$ 14.1 \\
    \hline feedforward & tax vectors & question & 54 & 20 $\pm$ 15.8 \\
    neural             &             & context  & 49 & 20 $\pm$ 15.8 \\
                       &             & statutes & 50 & 20 $\pm$ 15.8 \\
                       & word2vec    & question & 50 & 20 $\pm$ 15.8 \\
                       &             & context  & 50 & 20 $\pm$ 15.8 \\
                       &             & statutes & 50 & 20 $\pm$ 15.8 \\
    \hline feedforward & tax vectors & question & 51 & 20 $\pm$ 15.8 \\
    non-neural         &             & context  & 51 & 20 $\pm$ 15.8 \\
                       &             & statutes & 47 & 25 $\pm$ 17.1 \\
                       & word2vec    & question & 52 & 20 $\pm$ 15.8 \\
                       &             & context  & 51 & 20 $\pm$ 15.8 \\
                       &             & statutes & 53 & 20 $\pm$ 15.8 \\
  \end{tabular}
  \caption{Test set scores. We report the 90\% confidence interval. All confidence intervals for entailment round to 8.3\%.}
  \label{tab:results}
  \vspace{-25pt}
\end{table}

We use Adam \citep{kingma2014adam} with a linear warmup schedule for the learning rate. We freeze BERT's parameters, and experiment with unfreezing BERT's top layer. We select the final model based on early stopping with a random 10\% of the training examples reserved as a dev set. The best performing model for entailment and for numerical questions are selected separately, during a hyperparameter search around the recommended setting (batch size=32, learning rate=1e-5). To check for bias in our dataset, we drop either the statute, or the context and the statute, in which case we predict the answer from BERT's representation for ``[CLS]~+~$c$~+~[SEP]~+~$q$~+~[SEP]" or ``[CLS]~+~$q$~+~[SEP]", whichever is relevant.

\mysubsection{Feedforward models}
\label{subsec:feedforward}

We follow \citet{arora2016simple} to embed strings into vectors, with smoothing parameter equal to $10^{-3}$. We use either tax vectors described in Section \ref{sec:legal} or word2vec vectors \citep{mikolov2013distributed}. We estimate unigram counts from the corpus used to build the tax vectors, or the training set, whichever is relevant. For a given context $c$ and question or prompt $q$, we retrieve relevant subsection $s$ as above. Using \citet{arora2016simple}, $s$ is mapped to vector $v_s$, and $(c,q)$ to $v_{c+q}$. Let $r = [v_s, v_{q+c}, |v_s-v_{c+q}|, v_s\odot v_{c+q}]$ where $[a,b]$ is the concatenation of $a$ and $b$, $|.|$ is the element-wise absolute value, and $\odot$ is the element-wise product. The answer is predicted as $g(\theta_1 \cdot f(r))$ or $\mu + \sigma \theta_2 \cdot f(r)$, as above, where $f$ is a feed-forward neural network. We use batch normalization between each layer of the neural network \citep{ioffe2015batch}. As above, we perform ablation experiments, where we drop the statute, or the context and the statute, in which case $r$ is replaced by $v_{c+q}$ or $v_q$. We also experiment with $f$ being the identity function (no neural network). Training is otherwise done as above, but without the warmup schedule.

\mysubsection{Results}

We report the accuracy on the test set (in \%) in Table \ref{tab:results}. In our ablation experiments, ``question" models have  access to the question only, ``context" to the context and question, and ``statute" to the statutes, context and question. For entailment, we use a majority baseline. For the numerical questions, we find the constant that minimizes the hinge loss on the training set up to 2 digits: \$11,023. As a check, we swapped in the concatenation of the RTE datasets of \citet{dagan2005pascal,haim2006second,giampiccolo2007third,bentivogli2009fifth}, and achieved 73.6\% accuracy on the dev set with BERT, close to numbers reported in \citet{wang2019superglue}. BERT was trained on Wikipedia, which contains snippets of law text: see article \emph{United States Code} and links therefrom, especially \emph{Internal Revenue Code}. Overall, models perform comparably to the baseline, independent of the underlying method. Performance remains mostly unchanged when dropping the statutes or statutes and context, meaning that models are not utilizing the statutes. Adapting BERT or word vectors to the legal domain has no noticeable effect. Our results suggest that performance will not be improved through straightforward application of a large-scale language model, unlike it is on other datasets: \citet{raffel2019exploring} achieved 94.8\% accuracy on COPA \citep{roemmele2011choice} using a large-scale multitask Transformer model, and BERT provided a huge jump in performance on both SQuAD 2.0 \citep{rajpurkar2018know} (+8.2 F1) and SWAG \citep{zellers2018swag} (+27.1 percentage points accuracy) datasets as compared to predecessor models, pre-trained on smaller datasets.


Here, we focus on the creation of resources adapted to the legal domain, and on testing off-the-shelf and historical solutions. Future work will consider specialized reasoning models.

\mysection{Related work}
\label{sec:related}

There have been several efforts to translate law statutes into expert systems. Oracle Policy Automation has been used to formalize rules in a variety of contexts. TAXMAN \citep{mccarty1976reflections} focuses on corporate reorganization law, and is able to classify a case into three different legal types of reorganization, following a theorem-proving approach. \citet{sergot1986british} translate the major part of the British Nationality Act 1981 into around 150 rules in micro-Prolog, proving the suitability of Prolog logic to express and apply legislation. \citet{bench1987logic} further discuss knowledge representation issues. Closest to our work is \citet{sherman1987prolog}, who manually translated part of Canada's Income Tax Act into a Prolog program. To our knowledge, the projects cited did not include a dataset or task that the programs were applied to. Other works have similarly described the formalization of law statutes into rule-based systems \citep{hellawell1980computer,satoh2010proleg,fungwacharakorn2018legal,khan2016conversion}.

\citet{yoshioka2018overview} introduce a dataset of Japanese statute law and its English translation, together with questions collected from the Japanese bar exam. To tackle these two tasks, \citet{kim2019statute} investigate heuristic-based and machine learning-based methods. A similar dataset based on the Chinese bar exam was released by \citet{zhong2019jec}. Many papers explore case-based reasoning for law, with expert systems \citep{popp1974judith,vdl1983design}, human annotations \citep{bruninghaus2003predicting} or automatic annotations \citep{ashley2009automatically} as well as transformer-based methods \citep{rabelo2019combining}. Some datasets are concerned with very specific tasks, as in tagging in contracts \citep{chalkidis2017deep}, classifying clauses \citep{chalkidis2018obligation}, and classification of documents \citep{chalkidis2019large} or single paragraphs \citep{biagioli2005automatic}. \citet{ravichander2019question} have released a dataset of questions about privacy policies, elicited from turkers and answered by legal experts. \citet{saeidi2018interpretation} frame the task of statutory reasoning as a dialog between a user and a dialog agent. A single rule, with or without context, and a series of followup questions are needed to answer the original question. Contrary to our dataset, rules are isolated from the rest of the body of rules, and followup questions are part of the task.

\citet{clark2019f} describe a decades-long effort to answer science exam questions stated in natural language, based on descriptive knowledge stated in natural language. Their system relies on a variety of NLP and specialized reasoning techniques, with their most significant gains recently achieved via contextual language modeling.  This line of work is the most related in spirit to where we believe research in statutory reasoning should focus. An interesting contrast is that while scientific reasoning is based on understanding the physical world, which in theory can be informed by all manner of evidence beyond texts, legal reasoning is governed by human-made rules. The latter are true by virtue of being written down and agreed to, and are not discovered through evidence and a scientific process. Thus, statutory reasoning is an exceptionally pure instance of a reasoner needing to understand prescriptive language.

\citet{weston2015towards} introduced a set of \emph{prerequisite toy tasks} for AI systems, which require some amount of reasoning and common sense knowledge.  Contrary to the present work, the types of question in the train and test sets are highly related, and the vocabulary overlap is quite high.  Numeric reasoning appears in a variety of MR challenges, such as in DROP~\citep{dua2019drop}. 

Understanding procedural language -- knowledge needed to perform a task -- is related to the problem of understanding statutes, and so we  provide a brief description of some example investigations in that area.  \citet{zhang2012automatically} published a dataset of how-to instructions, with human annotations defining key attributes (actee, purpose...) and models to automatically extract the attributes. Similarly, \citet{chowdhury2020towards} describe a dataset of human-elicited procedural knowledge, and \citet{wambsganss2019mining} automatically detect repair instructions from posts on an automotive forum. \citet{branavan2012learning} employed text from an instruction manual to improve the performance of a game-playing agent.

\mysection{Conclusion}

We introduce a resource of law statutes, a dataset of hand-curated rules and cases in natural language, and a symbolic solver able to represent these rules and solve the challenge task. Our hand-built solver contrasts with our baselines based on current NLP approaches, even when we adapt them to the legal domain.

The intersection between NLP and the legal domain is a growing area of research \citep{rissland2003ai,ashley2009automatically,chalkidis2018obligation,kornilova-eidelman-2019-billsum,kim2019statute}, but with few large-scale systematic resources. Thus, in addition to the exciting challenge posed by statutory reasoning, we also intend this paper to be a contribution to legal-domain natural language processing.

Given the poor out-of-the box performance of otherwise very powerful models, this dataset, which is quite small compared to typical MR resources, raises the question of what the most promising direction of research would be. An important feature of statutory reasoning is the relative difficulty and expense in generating carefully constructed training data: legal texts are written for and by lawyers, who are cost-prohibitive to employ in bulk.  This is unlike most instances of MR where everyday texts can be annotated through crowdsourcing services.  There are at least three strategies open to the community: automatic extraction of knowledge graphs from text with the same accuracy as we did for our Prolog solver \citep{viet2017convamr};  improvements in MR to be significantly more data efficient in training; or new mechanisms for the efficient creation of training data based on pre-existing legal cases.

Going forward, we hope our resource provides both (1) a benchmark for a challenging aspect of natural legal language processing as well as for machine reasoning, and (2) legal-domain NLP models useful for the research community.

\bibliographystyle{ACM-Reference-Format}
\bibliography{sample-base}

\end{document}